%% file: acl_latex.tex
\documentclass[11pt]{article}

\usepackage[preprint]{acl}

\usepackage{times}
\usepackage{latexsym}

\usepackage[T1]{fontenc}

\usepackage[utf8]{inputenc}

\usepackage{microtype}

\usepackage{inconsolata}

\usepackage{graphicx}

\usepackage{breqn}

\usepackage{algorithm}
\usepackage{algorithmic}

\usepackage{booktabs}
\usepackage{multirow}
\usepackage{siunitx}
\usepackage{amsmath}
\usepackage[table]{xcolor}
\usepackage[dvipsnames]{xcolor}
\usepackage{makecell}
\newcommand{\highlightbest}[2]{%
    \ifdim#1pt=#2pt \bfseries\color{RoyalBlue}#1\else#1\fi%
}
\definecolor{Gray}{gray}{0.9}
\definecolor{LightCyan}{rgb}{0.88,1,1}
\definecolor{highlightcolor}{rgb}{0.9, 0.9, 1} 

\usepackage{amssymb}
\usepackage{pifont}

\usepackage{hyperref}

\usepackage{stfloats}

\title{DARO: Difficulty-Aware Reweighting Policy Optimization}

\author{
    \textbf{Jingyu Zhou}\textsuperscript{1,*},
    \textbf{Lu Ma}\textsuperscript{2,*},
    \textbf{Hao Liang}\textsuperscript{2},
    \textbf{Chengyu Shen}\textsuperscript{2},
    \textbf{Bin Cui}\textsuperscript{2},
    \textbf{Wentao Zhang}\textsuperscript{2}
\\
\\
    \textsuperscript{*}Equal Contribution
\\
    \textsuperscript{1}Shanghai Jiao Tong University,
    \textsuperscript{2}Peking University
}

\begin{document}
\maketitle

\begin{figure*}
  \includegraphics[width=1.0\textwidth]{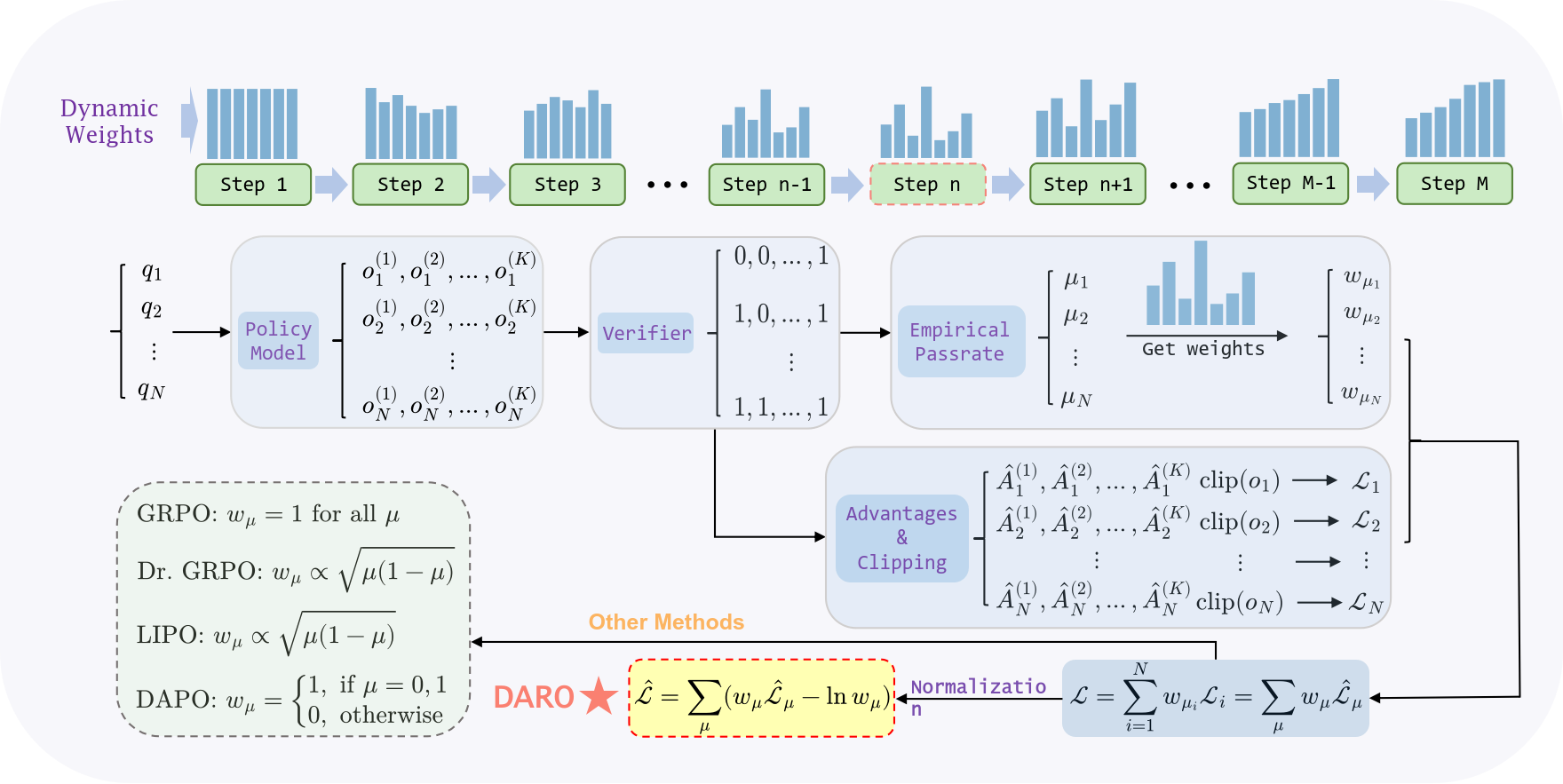}
  \label{fig:pipeline}
  \captionof{figure}{\textbf{The Dynamic Adaptive Reweighting Policy Optimization(DARO) Framework.} This figure illustrates the training pipeline where, for each prompt, an empirical pass rate ($\mu_i$) and a base loss ($\mathcal{L}_i$) are computed using verifier rewards. Unlike methods that use static weighting functions based on $\mu$ (e.g., GRPO, DAPO, Dr. GRPO and LIPO), DAPO treats the weights ($w_\mu$) as optimizable parameters. This approach allows the weights to adapt dynamically to the model's state, as visualized by the evolving weight distributions across training steps (top).}
  \label{fig:experiments}
\end{figure*}

\begin{abstract}
Recent advances in large language models (LLMs) have shown that reasoning ability can be significantly enhanced through Reinforcement Learning with Verifiable Rewards (RLVR). Group Relative Policy Optimization (GRPO) has emerged as the de facto approach for RLVR, inspiring numerous variants. However, our mathematical analysis reveals that these methods are fundamentally weighted variations of GRPO. We provide a unified view, demonstrating that their reliance on static or overly simplistic weighting schemes tied to sample difficulty prevents adaptation to a model's evolving capabilities. This creates a significant loss scale issue, where training disproportionately focuses on certain difficulty levels at the expense of others, hindering overall performance. To address these limitations, we introduce \textbf{Difficulty-Aware Reweighting Policy Optimization (DARO)}, a method that dynamically adjusts the loss contribution of each difficulty group based on the model's learning state. Extensive experiments on Qwen2.5-Math-1.5B, Qwen2.5-Math-7B, and Llama3.1-8B show that DARO outperforms four leading baselines across six math benchmarks, achieving significantly faster convergence and superior final performance.
\end{abstract}

\input{sections/introduction}

\input{sections/related_works}

\input{sections/motivation}

\input{sections/method}

\input{sections/experiment}

\input{sections/conclusion}

\input{sections/limitations}



\bibliography{acl_latex}

\input{sections/appendix}

\end{document}

%% file: sections/introduction.tex
\section{Introduction}

While LLMs have demonstrated remarkable proficiency across a wide spectrum of natural language tasks, enhancing their reasoning abilities remains a critical frontier in artificial intelligence research ~\cite{ke2025survey,sun2025survey,zhang2025survey,zhang2025100}. A leading strategy for this purpose is Reinforcement Learning with Verifiable Rewards (RLVR), a paradigm that fine-tunes LLMs using binary feedback from verifiable answers. This approach enables models to autonomously explore a vast solution space and learn from their own generated solutions, thereby transcending the limitations of supervised fine-tuning on static datasets.

The success of RLVR is largely driven by algorithms derived from Proximal Policy Optimization (PPO) ~\cite{ppo} and Group Relative Policy Optimization (GRPO) ~\cite{grpo}. GRPO, in particular, has inspired several prominent variants, including DAPO~\cite{yu2025dapo}, Dr. GRPO~\cite{dr.grpo}, and Lite PPO (LIPO)~\cite{lipo}. These methods achieve significant success by optimizing a model's policy to maximize verifiable rewards, thereby elevating its performance on complex mathematical benchmarks. By sampling multiple responses for a given prompt and normalizing rewards within that group, this framework has become a cornerstone of modern LLM post-training for advanced reasoning.


Despite their success, our investigation reveals that these algorithms share a critical, unaddressed limitation. To analyze this, we introduce a unified loss form that frames existing methods as weighted variations of GRPO. Our analysis shows that their intrinsic weighting schemes, which are tied to sample difficulty (i.e., empirical passrate), are either static or overly simplistic, failing to adapt to the training progress and different model types. This flaw creates what we term the \textit{loss scale issue}: an imbalance where the training objective disproportionately focuses on samples at certain difficulty level.

To resolve this challenge, we propose \textbf{D}ifficulty-\textbf{A}ware \textbf{R}eweighting Policy \textbf{O}ptimization (\textbf{DARO}), an adaptive algorithm that dynamically modulates the loss contribution of each difficulty group based on the model's learning progress. In contrast to methods that rely on fixed heuristics, DARO intelligently recalibrates difficulty weights to ensure that the model both consolidates existing knowledge and efficiently explores new, challenging problems. This approach alleviates the loss scale issue, fostering a more stable and effective learning trajectory. Comprehensive experiments on six math reasoning benchmarks across three base models demonstrate that DARO outperforms GRPO, Dr. GRPO, DAPO, and LIPO, achieving both significantly faster convergence and superior final performance.
The primary contributions of this paper are as follows:

\begin{itemize}
    \item The identification and formalization of the "\textbf{loss scale issue}", a inherent critical flaw in current RLVR algorithms, supported by both theoretical and empirical evidence of its detrimental impact.
    \item The introduction of \textbf{DARO}, a novel, dynamic weighting algorithm designed to resolve loss scale issue by adapting the training loss to the model's learning state.
    \item Extensive experiments show DARO achieves state-of-the-art performance on mathematical reasoning benchmarks, outperforming GRPO, LIPO, Dr. GRPO and DAPO with a much faster convergence speed.
\end{itemize}

%% file: sections/related_works.tex
\section{Related Works}
\subsection{Emergent Reasoning via RL}
RL has recently emerged as a promising approach for enhancing the reasoning capabilities of LLMs. Early breakthroughs, such as OpenAI o-series~\cite{openai-o1,openai-o3}, DeepSeek-R1~\cite{deepseek_r1,marjanovic2025deepseek}, Gemini 2.5 ~\cite{comanici2025gemini,huang2025gemini}, and kimi k-series~\cite{kimi1.5,kimik2}, have demonstrated that applying RLVR can lead to significant improvements in complex reasoning tasks. Building upon these foundations, Light-R1~\cite{Logic-rl} incorporates curriculum learning to refine LLM's reasoning skills, while Logic-RL~\cite{Logic-rl} adopts logic-driven reward functions to improve general reasoning ability. Deepscaler~\cite{Deepscaler} trains LLMs with progressively longer contexts as their performance improves.  ~\citep{yue2025does, yan2025learning, ma2025learning} focus on introducing and addressing the concern that RLVR does not impart fundamentally new reasoning abilities to LLMs.

\subsection{Improved Policy Optimization for RLVR}
Following the success of Deepseek-R1 ~\cite{deepseek_r1}, GRPO ~\cite{grpo} has become a de facto standard for RLVR training. Its core principle is to estimate the advantage directly from the reward scores of multiple sampled solutions, thereby obviating the need for a separate value model. The original GRPO framework has inspired several refinements aimed at improving its stability and performance. To address the biases in GRPO, Dr.GRPO ~\cite{dr.grpo} simplifies the original algorithm by removing the standard deviation from the advantage calculation and excluding token-level normalization. Another variant, LIPO ~\cite{lipo}, focuses on achieving superior performance through a more sophisticated advantage normalization technique, which incorporates both group-level mean and batch-level standard deviation, complemented by token-level loss aggregation. Along similar lines, DAPO~\cite{yu2025dapo} uses clip-higher to prevent entropy collapse, dynamic sampling for improved efficiency, and token-level policy gradient loss with overlong reward shaping to stabilize training. We find that all these variants rely on a unified loss formulation with various weighting factors, but DARO distinguishes itself by proposing better, learnable weighting factors, which allows it to achieve superior performance.

%% file: sections/motivation.tex
\section{Preliminary \& Theoretical Guarantees}
\label{sec:analysis}

\subsection{A Unified Loss Formulation}


To analyze the internal mechanisms of existing policy optimization algorithms, we begin with the GRPO loss function, simplified for the case of binary $\{0, 1\}$ rewards. Based on recent studies showing the KL divergence term to be non-essential ~\cite{hu2025open,he2025skywork,yu2025dapo} and adopting a more general token-mean loss aggregation ~\cite{yu2025dapo,lipo}, the simplified GRPO loss can be expressed as:
\begin{equation}
    \begin{split}
    &\mathcal{L}_{GRPO}
    = \\ & \mathbf{E}_{q\sim \mathcal{Q},\{ o_{i} \}_{i=1}^{K}\sim \pi_{\theta_{old}}(\cdot|q)} \left[\sum_{i=1}^{K} \frac{\lvert o_{i} \rvert}{L} f(A_i, o_{i})\right], \\
    &f(A,o) = \begin{cases}
        \min \{ r(\theta )A,(1+\epsilon)A \},  & A>0 \\
 \\
\max_{}\{ r(\theta)A, (1-\epsilon)A \}, & A<0
    \end{cases}
    \end{split}
\end{equation}

In addition, $\mathcal{Q}$ is a distribution of prompts, $o_{i}$ is the $i$-th response generated by the LLM $\pi_{\theta_{old}}$ from prompt $q$. The term $\lvert o_{i} \rvert$ represents the length of the response, $L=\sum_{i=1}^L\lvert o_i\rvert$ is the total response length across $\mathcal{Q}$, $r_i$ is the reward of the response. The advantage of $o_{i}$ is calculated as $A_{i}=\frac{r_{i}-\mu}{\sigma}$, where $\mu$ and $\sigma$ are the mean and standard deviation of rewards across the $K$ responses from $q$. Notably, $\mu$ also represents the empirical pass rate for $q$. For simplified analysis, we denote $L=\sum_{i=1}^{K}\lvert o_{i} \rvert$ as the total length of all responses and $r(\theta)=\pi_\theta(o)/\pi_{\theta_{old}}(o)$. This objective can be generalized by replacing the implicit constant $1$ with a variable weight $w_i$, yielding the following form for the base loss:
\begin{equation}
    \begin{split}
    \mathcal{L}_{BASE}=& \mathbf{E}_{q\sim \mathcal{Q},\{ o_{i} \}_{i=1}^{K}\sim \pi_{\theta_{old}}(\cdot|q)} \\
    &\left[\sum_{i=1}^{K} w_i\cdot \frac{\lvert o_{i} \rvert}{L} f(A_i, o_{i})\right].
    \end{split}
    \label{eq:base_loss}  
\end{equation}

\begin{table}[t]
    \centering
    \caption{Comparison of Weighting Factors in Different RLVR Algorithms.}
    \label{tab:weight_comparison}
    \begin{tabular}{l l}
        \toprule
        \textbf{Algorithm} & \textbf{Weight} \\
        \midrule
        GRPO               & $w_i = 1$ \\
        
        DAPO               & $w_i=\mathbb{I}(0<\mu<1)$ \\
        
        LIPO               & $w_{i} \propto \sqrt{\mu(1-\mu)}$ \\
        
        Dr. GRPO           & $w_i \propto \sqrt{\mu(1-\mu)}$ \\
        \bottomrule
    \end{tabular}
\end{table}

DAPO, LIPO, and Dr.GRPO introduce different weighting factors $w_i$ to $\hat{\mathcal{L}}_{BASE}$, as summarized in Table~\ref{tab:weight_comparison} and detailed in Appendix \hyperref[appendix:obj]{A.2}. DAPO treats all samples with an empirical pass rate $\mu\in(0,1)$ identically, which fails to capture nuanced differences among samples of varying difficulty, potentially limiting training efficiency. LIPO uses the batch-level standard deviation $\hat{\sigma}$ instead of the group-level deviation $\sigma$, while Dr. GRPO removes both the standard deviation $\sigma$ and the reduction factor $L$. Consequently, the weight for both Dr.RPO and LIPO can be expressed as $w(\mu) \propto \sqrt{\mu(1-\mu)}$ (see derivation in Appendix~\ref{appendix:adv}). This function peaks at $\mu=0.5$ and approaches zero as $\mu$ nears 0 or 1, revealing an intrinsic characteristic of these methods: they prioritize medium-difficulty samples while downweighting both easy ($\mu \to 1$) and extremely difficult ($\mu \to 0$) samples. This strategic focus creates a risk of catastrophic forgetting, where the model may lose knowledge from simple samples while focusing on medium-difficulty ones.

\subsection{The Loss Scale Issue}
\label{sec:loss_scale_issue}

\begin{figure*}[t!]
  \includegraphics[width=\linewidth]{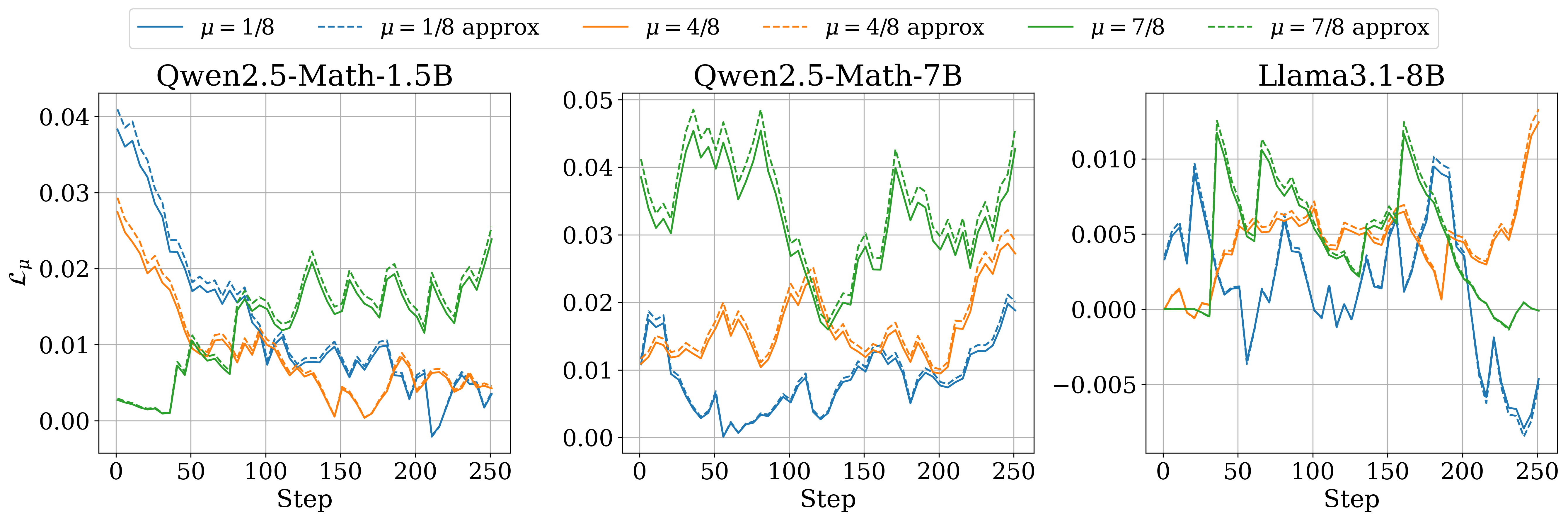}
  \caption{\textbf{Loss Scale Issue} observed in the GRPO training process of three base models, exponentially smoothed with $\alpha=0.1$. The solid line represents the true loss value, while the dashed line represents the loss value approximately obtained through equation~\ref{eq:theorem_1}. Detailed hyperparameters are shown in Section~\ref{sec:settings}.}
  \label{fig:loss_scale}
\end{figure*}

The unified loss function presented above reveals that existing algorithms rely on static, often simplistic weighting schemes determined by the empirical pass rate $\mu$. This raises the critical question of whether such pre-defined strategies are optimal. To investigate the contribution of samples at different difficulty levels, we propose reframing the RL process as a multitask learning problem. In this framework, each empirical pass rate $\mu = k/K$, where $k \in \{0,1, \dots, K\}$ is the number of correct responses, corresponds to a distinct task. This framework enables us to examine the behavior of the base loss (Equation~\ref{eq:base_loss}) on groups of prompts $G_\mu$, where each group comprises prompts sharing the same pass rate $\mu = k/K$. To facilitate further analysis, we separate positive ($r=1$) and negative ($r=0$) responses within $G_\mu$. Denoting the $i$-th responce of the $j$-th sample as $o_i^{(j)}$, we define the losses for positive and negative samples as:

\begin{equation}
\begin{split}
    &\mathcal{L}_{BASE}(G_\mu)=-\frac{1}{nK} [\mathcal{L}_{+}(G_\mu) + \mathcal{L}_{-}(G_\mu)] \\
    &\text{where }\mathcal{L}_{\pm}=\sum_{o_i^{(j)} \text{ is pos/neg}} \frac{\lvert o_i^{(j)}\rvert}{L} f(A_{\pm}, o_{i}^{(j)}).
\end{split}
\end{equation}

A direct analysis of the base loss is complicated by the clip term $f(o_{i})$. However, since $f(A,\cdot)$ is bounded within the interval $[0,(1+\epsilon)A]$ when $A>0$ or $[(1-\epsilon)A,0]$ when $A<0$, we can approximate these losses using Hoeffding's inequality. This approximation introduces a bounded deviation that is negligible for our purposes. See Appendix~\ref{appendix:lemma1} for the full derivation. Specifically, the base loss can be approximated as:

\begin{equation}
    \begin{split}
&\mathcal{L}_{BASE}(G_{k})
\approx\\& \frac{\sum_{o\text{ is pos}} \lvert o\rvert - \sum_{o\text{ is neg}} \lvert o\rvert}{L} \sqrt{\mu(1-\mu)}
    \label{eq:theorem_1}        
    \end{split}
\end{equation}



\begin{figure}[th]
\includegraphics[width=\columnwidth]{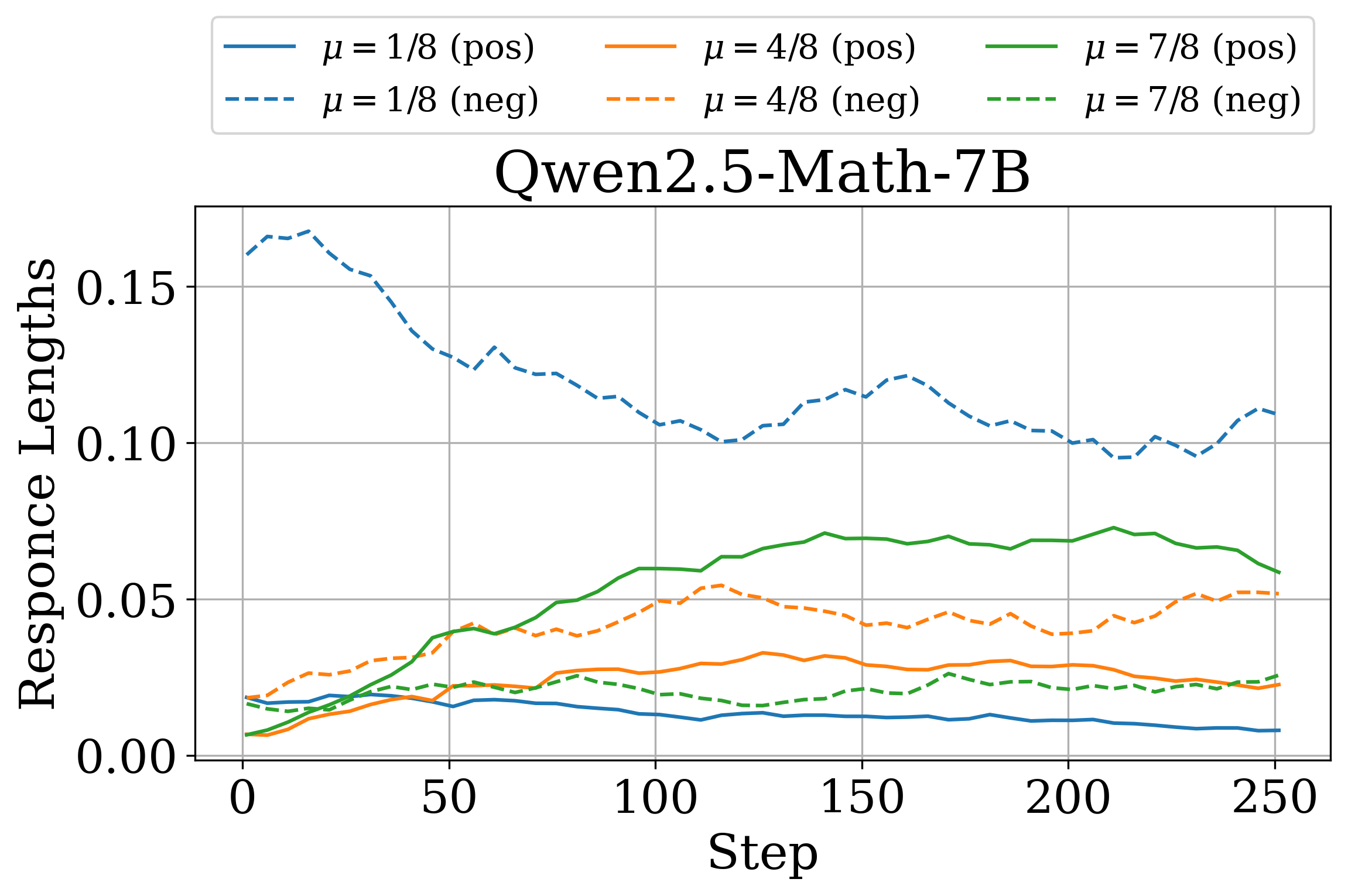}
  \caption{\textbf{Normalized Response Lengths} of Qwen2.5-Math-7B model in the GRPO training process. The responce lengths are normalized by dividing the sum of responce lengths across the batch $B$ at each step. The solid and dashed lines represents the responce lengths of positive ($r=1$) and negative ($r=0)$ samples, respectively.}
\label{fig:responce_length_mat7b}
\end{figure}

As mentioned in \cite{lipo}, responce length is often correlated with the difficulty level of samples and we conduct experiments to empirically verify this assertion. The result of Qwen2.5-Math-7B is shown in Figure~\ref{fig:responce_length_mat7b}, and see Appendix~\ref{appendix:detailed_results} for more detailed results. These results further confirm the correlation. Therefore, Equation~\ref{eq:theorem_1} implies that the magnitude of the loss is strongly linked to the sample's empirical pass rate $\mu$. To validate the theoretical analysis, we conduct experiments on Qwen2.5-Math-1.5B, Qwen2.5-Math-7B and Llama3.1-8B, tracking the mean loss of sub-batches of samples with different empirical passrates throughout training.

As shown in Figure~\ref{fig:loss_scale}, it is clear that the loss scale issue indeed exists and the approximation of Equation~\ref{eq:theorem_1} is reasonable. In the whole training process, \textbf{the loss varies with model types, values of $\mu$ and the progress of training, and there is always one pass rate $\mu$ that $\mathcal{L}_\mu$ is much higher compared to others}. This indicates that the model is forced to focus on samples at a certain difficulty level while relatively ignoring samples at the remaining levels. This disparity of loss scale, which we term the "\textit{loss scale issue}", causes the model to either rote-learn existing knowledge or pay too much attention to those extremely difficult problems. Consequently, this disrupts the balance between exploration and exploitation, potentially impeding both the efficiency and the final effectiveness of training. Furthermore, for methods such as DAPO, their binary weighting strategy can not eliminate the issue for the positive ($r=1$) samples. For methods including Dr.GRPO and LIPO, their pre-defined weighting strategies can not adapt to the varying losses, which are even possible to exacerbate the loss scale issue, causing a more negative impacts on training efficiency and the final results.

%% file: sections/method.tex
\section{Method}

\begin{algorithm*}[t]
\caption{\textbf{D}ifficulty-\textbf{A}ware \textbf{R}eweighting Policy \textbf{O}ptimization (\textbf{DARO})}
\label{alg:adaptive_weighting}
\begin{algorithmic}[1]
\STATE \textbf{Initialize:} Model parameters $\theta$, adaptive weight parameters $\{w_\mu\}$, reward model $R$, dataset $\mathcal{Q}$, hyperparameter $C$.

\FOR{step $=1,2,\dots,$}
    \STATE Sample a batch $B = \{q_1, \dots, q_N\}$ from $\mathcal{Q}$.
    \STATE Update the old policy $\pi_{\theta_{\text{old}}} \leftarrow \pi_\theta$
    \STATE Sample $K$ outputs, $\{o_1,o_2,\dots,o_K\}$, for each quest $q$.
    \STATE Compute rewards for each output $o_i$ using $R$; empirical pass rate $\mu_q$ for each query $q$; total response length $L$ for queries where $0 < \mu_q < 1$.
    \STATE Compute $\mathcal{L}_\mu$ for each possible $\mu$
    \STATE Calculate the total loss $\mathcal{L}_{\text{total}} \leftarrow \sum_\mu (w_\mu\mathcal{L}_\mu-C\ln w_\mu)$
    \STATE Compute gradients of $\mathcal{L}_{total}$ with respect to model parameters $\theta$ and weight parameters $\{w_\mu\}$ then update $\theta$ and $\{w_\mu\}$.
\ENDFOR
\end{algorithmic}
\end{algorithm*}

Motivated by the aforementioned findings, we aim to develop a weighting mechanism that dynamically adapts to the model's evolving capabilities and alleviate the loss scale issue. To achieve this, we adopt a multitask learning framework where groups of samples with varying empirical pass rates are treated as distinct tasks. We then assign a learnable weight parameter, $w_\mu$, where $\mu=k/K$, to each task (i.e., $k \in \{1, \dots, K-1\}$).


Let $\mathcal{L}_{\mu}$ be the loss for the group of samples with an empirical pass rate of $\mu=k/K$. The goal is to find an optimal set of weights $\{w_\mu\}$ to define a total weight loss, $\mathcal{L} = \sum_{\mu} w_\mu \mathcal{L}_\mu$, which promotes more efficient training. To address the loss scale issue, the weights should ideally be inversely proportional to the original loss: $w_\mu \propto \mathcal{L}_\mu^{-1}$. This relationship ensures that the weighted loss for each task, $w_\mu\mathcal{L}_\mu$, is approximately equal.

To achieve this inverse proportionality during optimization and prevent the weights from converging to a trivial solution (e.g., all weights becoming zero), we introduce a regularization term, $N(w_\mu)$, for each task's weight. The regularized weighted loss for each task $\mu$ is then defined as:
\begin{equation}
\hat{\mathcal{L}}_{\mu}=w_{\mu}\mathcal{L}_{\mu} + N(w_{\mu}).
\end{equation}

The model parameters $\theta$ and the weights $w_\mu$ are updated at each optimization step. To find the optimal weight for a given task, we set the partial derivative of the total loss, $\mathcal{L}=\sum_{\mu}\hat{\mathcal{L}}_{\mu}$, with respect to $w_\mu$ and setting it to zero:
\begin{equation}
\nabla_{w_{\mu}}\hat{\mathcal{L}}=\mathcal{L}_{\mu} + N'(w_{\mu}) = 0.
\end{equation}

To satisfy the ideal condition $w_{\mu}=C\mathcal{L}_{\mu}^{-1}$ (for a positive constant $C$), we substitute $\mathcal{L}_\mu = Cw_\mu^{-1}$ into the optimality condition $mathcal{L}_{\mu} + N'(w_{\mu}) = 0$. This implies that the regularizer's derivative must be $N'(w_{\mu}) = -Cw_\mu^{-1}$. Integrating this expression yields the regularization term. For simplicity, we set the constant $C$ to 1.

This derivation leads to our final loss function. Furthermore, following the effective practice of DAPO ~\cite{yu2025dapo}, we set the weights for samples with $\mu=0,1$ to zero, as their advantage is zero and they do not contribute to the policy gradient. The final loss is therefore defined as:
\begin{equation}
    \mathcal{L} = \sum_{\mu\ne0,1} (w_{\mu}\mathcal{L}_{\mu}-\ln w_{\mu}),
\end{equation}
where $\mathcal{L}_\mu$ represents the loss for samples with an empirical pass rate $\mu$. The weights $w_\mu$ are treated as trainable parameters and are optimized jointly with the model parameters $\theta$. Our proposed DARO is detailed in Algorithm~\ref{alg:adaptive_weighting}.

%% file: sections/experiment.tex
\begin{table*}[!t]
    \centering
    \caption{Overall accuracy (\%) on six math benchmarks based on Llama-3.1-8B, Qwen2.5-Math-1.5B, and Qwen2.5-Math-7B. The best result in each column for a given model is highlighted in \textbf{\color{RoyalBlue}blue} except the "Average" column with \textbf{\color{WildStrawberry}red}. When the value in \textbf{DARO} row is not the best, it is underlined if sub-optimal.}
    \label{tab:main_results}
    \sisetup{table-format=2.1, detect-weight} 
    \resizebox{1.0\linewidth}{!}{%
    \begin{tabular}{p{42pt} S[table-format=2.1] S[table-format=2.1] S[table-format=2.1] S[table-format=1.1] S[table-format=2.1] S[table-format=2.1] S[table-format=2.1] S[table-format=2.1, detect-weight, table-text-alignment=center, table-number-alignment=center]}
        \toprule
        \cmidrule(lr){2-9}
         & {\bf \makecell{MATH500\\mean@1}} & {\bf \makecell{AMC23\\mean@32}} & {\bf \makecell{AIME24\\mean@32}} & {\bf \makecell{AIME25\\mean@32}} & {\bf \makecell{Olympiad\\mean@1}} & {\bf \makecell{Minerva\\mean@1}} & {\bf \makecell{GSM8K\\mean@1}} & {\textbf{Average}} \\
        \midrule
        \multicolumn{9}{c}{\textit{Llama-3.1-8B}} \\
        \midrule
        GRPO    & 25.2 & 12.1 & \highlightbest{2.1}{2.1}  & 0.2 & 7.3  & 17.6 & 66.3 & 18.7 \\
        LIPO    & 23.0 & 10.8 & 0.0  & 0.0 & 5.2  & 17.3 & 60.0 & 16.6 \\
        Dr.GRPO & 20.4 & 10.5 & 0.1  & \highlightbest{0.9}{0.9} & 8.5  & 21.7 & 62.5 & 17.8 \\
        DAPO    & 29.8 & 12.3 & 1.0  & 0.4 & 8.7  & 21.0 & 68.8 & 20.3 \\
        \rowcolor{Gray}
        DARO   & \highlightbest{30.2}{30.2} & \highlightbest{12.9}{12.9} & \underline{1.0}  & 0.0 & \highlightbest{10.1}{10.1} & \highlightbest{22.4}{22.4} & \highlightbest{73.0}{73.0} & \bfseries {\color{WildStrawberry}{21.4}} \\
        \midrule
        \multicolumn{9}{c}{\textit{Qwen2.5-Math-1.5B}} \\
        \midrule
        GRPO    & 68.6 & 45.8 & 8.8  & 7.3 & 34.8 & \highlightbest{31.3}{31.3} & 80.4 & 39.6 \\
        LIPO    & 70.6 & \highlightbest{48.0}{48.0} & 10.8 & 7.7 & 34.0 & 27.6 & 80.4 & 39.9 \\
        Dr.GRPO & 70.0 & 45.3 & \highlightbest{11.5}{11.5} & 6.8 & 34.8 & 29.0 & 79.5 & 39.6 \\
        DAPO    & 67.4 & 46.3 & 9.7  & \highlightbest{8.0}{8.0} & \highlightbest{38.2}{38.2} & 28.7 & \highlightbest{82.6}{82.6} & 40.1 \\
        \rowcolor{Gray}
        DARO     & \highlightbest{73.8}{73.8} & \underline{46.6} & 10.4 & 5.7 & \underline{36.1} & \underline{29.8} & \underline{81.6} & \bfseries {\color{WildStrawberry}{40.6}} \\
        \midrule
        \multicolumn{9}{c}{\textit{Qwen2.5-Math-7B}} \\
        \midrule
        GRPO    & \highlightbest{84.0}{84.0} & 61.7 & 17.4 & 12.2 & 46.1 & 37.5 & 87.2 & 49.4 \\
        LIPO    & 79.6 & 61.5 & \highlightbest{20.5}{20.5} & 11.9 & 45.0 & 37.5 & 89.3 & 49.3 \\
        Dr.GRPO & 81.0 & 59.7 & 18.3 & 13.2 & 43.5 & \highlightbest{39.0}{39.0} & 85.7 & 48.6 \\
        DAPO    & 80.2 & 57.6 & 15.8 & 10.4 & 44.1 & 36.8 & 88.2 & 48.4 \\
        \rowcolor{Gray}
        DARO     & \underline{81.8} & \highlightbest{65.4}{65.4} & 17.6 & \highlightbest{14.6}{14.6} & \highlightbest{46.8}{46.8} & \underline{38.2} & \highlightbest{91.2}{91.2} & \bfseries {\color{WildStrawberry}{50.8}} \\
        \bottomrule
        \label{tab:best_acc}
    \end{tabular}%
    }
\end{table*}

\section{Experiments}

\begin{table*}[t]
    \centering
    \caption{Ablation study on DARO based on Llama-3.1-8B, Qwen2.5-Math-1.5B, and Qwen2.5-Math-7B. The results indicate that \textit{Dynamic Weights} strategy substantially improves the final average performance.}
    \resizebox{1.0\textwidth}{!}{
    \begin{tabular}{l c c c c c c c c}
         \toprule
         & \makecell{Token\\Mean} & \makecell{Clip\\Higher} & \makecell{Dynamic\\Sampling} &
         \makecell{Static\\Weights} &
         \makecell{Dynamic\\Weights} & {\bf \makecell{Qwen2.5\\Math 7B}} & {\bf \makecell{Qwen2.5\\Math 1.5B}} & {\bf \makecell{Llama3.1\\8B}} \\
         \midrule
         GRPO & \checkmark & \checkmark & \ding{55} & \ding{55} & \ding{55} & 49.4 (\textbf{+0.0}) & 39.6 (\textbf{+0.0}) & 18.7 (\textbf{+0.0}) \\
         DAPO & \checkmark & \checkmark & \checkmark & \ding{55} & \ding{55} & 48.4 ({\color{LimeGreen}\textbf{-1.0}}) & 40.1 (\textbf{+0.5}) & 20.3 (\textbf{+1.6}) \\
         Dr.GRPO & \checkmark & \checkmark & \ding{55} & \checkmark & \ding{55} & 48.6 ({\bf\color{LimeGreen}-0.8}) & 39.6 ({\bf+0.0}) & 17.8 ({\bf\color{LimeGreen}-0.9}) \\
         LIPO & \checkmark & \checkmark & \ding{55} & \checkmark & \ding{55} & 49.3 ({\bf\color{LimeGreen}-0.1}) & 39.9 (\textbf{+0.3}) & 16.6 ({\bf\color{LimeGreen}-2.1}) \\
         \rowcolor{Gray}
         \textbf{DARO} & \checkmark & \checkmark & \checkmark & \ding{55} & \checkmark & 50.8 ({\color{WildStrawberry}\textbf{+1.4}}) & 40.6 ({\color{WildStrawberry}\textbf{+1.0}}) & 21.4 ({\color{WildStrawberry}\textbf{+2.7}}) \\
         \bottomrule
    \end{tabular}
    }
    \label{tab:grpo_dapo_dar}
\end{table*}

\subsection{Settings}
\label{sec:settings}

\textbf{Models.} We evaluat the proposed DARO algorithm and four baselines, GRPO, DAPO, LIPO and Dr.GRPO, on three models, Qwen2.5-Math-1.5B, Qwen2.5-Math-7B and Llama-3.1-8B. Since the maximum sequence length of $4096$ for Qwen2.5-Math-1.5B and Qwen2.5-Math-7B is insufficient for complex reasoning tasks, we expanded it to $16384$. This is achieved increasing $\theta_{\text{ROPE}}$ from $10000$ to $40000$.

\noindent \textbf{Datasets.} We conduct training on a subset of OpenR1-Math-220k \cite{openr1}, filtered by LUFFY \cite{luffy}, resulting in a total of 45,000 prompts, denoted as OpenR1-45k. In addition, we filter this subset to create a simpler sub-subset, OpenR1-easy, containing 11,000 prompts, which is special for training Llama3.1-8B to accommodate its weaker reasoning capabilities. For evaluation, we utilize the MATH500 \cite{math500}, gsm8k \cite{gsm8k}, Minerva \cite{minverva}, OlympiadBench \cite{he2024olympiadbench}, AMC23, AIME24 and AIME25 \cite{numina_math_datasets}. These datasets range in difficulty from elementary (e.g., GSM8K) to highly challenging (e.g., AIME25), thereby forming a comprehensive evaluation suite that spans a wide spectrum of difficulty levels. To ensure the stability of evaluation, we report the mean@32 accuracy for AMC23, AIME24, and AIME25.

\noindent \textbf{Hyperparameters.} To ensure consistency across all experiments, we employ token-mean loss aggregation and clip-higher strategy. The lower and higher clipping ranges set to $\epsilon_{\text{low}}=0.2$ and $\epsilon_{\text{high}}=0.28$. The maximum prompt length is set to $1024$ and the maximum responce length is $8192$. The training batch size is $128$ while the mini batch size is $64$. We use the AdamW \cite{adamw} optimizer with a constant learning rate of $1\mathrm{e}{-6}$ for model parameters and $1\mathrm{e}{-3}$ for the dynamic weights. At each step, $8$ responses are generated for each sample. We also remove all KL terms. Specifically, for DAPO and DARO, the generation batch size is $384$ and samples with pass rate of $0$ or $1$ are removed from the batch. The overlong buffer trick is not used in DAPO. Models are trained for a total of $300$ steps. More detailed parameters are provided at Appendix~\ref{appendix:detailed_params}. 

\subsection{Main Results}


\begin{figure*}[t]
  \includegraphics[width=\textwidth]{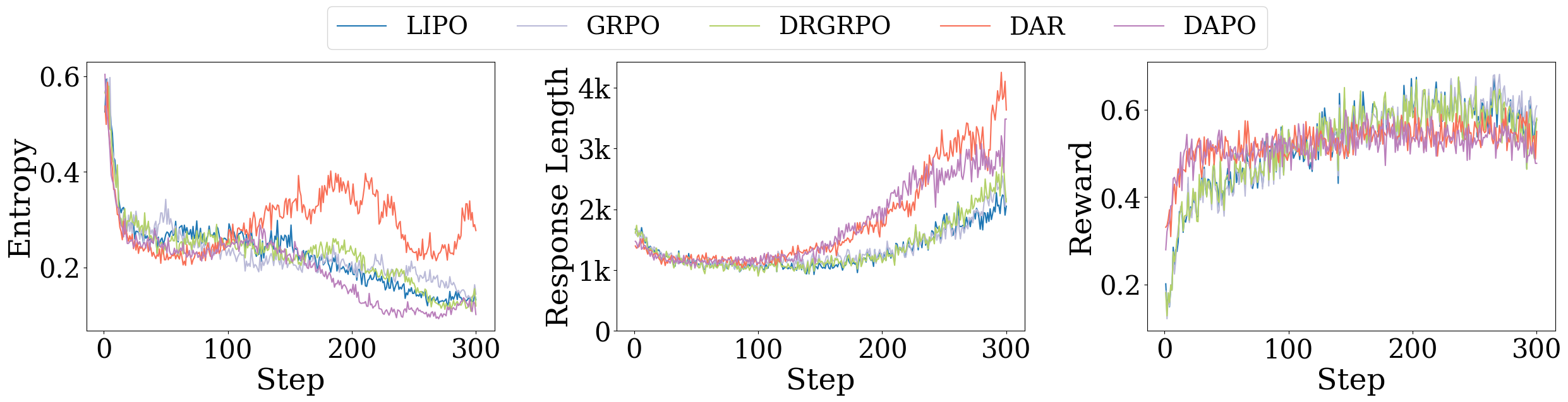}
  \caption{Entropy, responce length and train reward dynamic of Qwen2.5-Math-7B model throughout training.}
  \label{fig:train_dynamic}
\end{figure*}

As shown in Table~\ref{tab:best_acc}, our proposed DARO method consistently outperforms all competing methods, i.e. GRPO, LIPO, Dr.GRPO, and DAPO, across all three base models. It achieves the highest average accuracy on Llama-3.1-8B (21.4\%), Qwen2.5-Math-1.5B (40.6\%), and Qwen2.5-Math-7B (50.8\%). Furthermore, DARO maintains optimal or near-optimal performance on each individual benchmark, securing the top score in the majority of test cases. For instance, on the Llama-3.1-8B model, it wins 5 out of 7 benchmarks. Even when not the absolute best, its performance remains highly competitive, as indicated by the underlined sub-optimal results. These findings strongly indicate that DARO effectively enhances mathematical reasoning capabilities through its difficulty-aware reweighting strategy.

Our primary innovation, the Dynamic Weights strategy, proves to be highly effective, as demonstrated by the ablation study in Table~\ref{tab:grpo_dapo_dar}. When this component is added, our full DARO model consistently and substantially outperforms all other variants across three different base models. Specifically, DARO achieves performance gains of $+1.4$, $+1.0$, and $+2.7$ over the GRPO baseline on Qwen2.5-Math-7B, Qwen2.5-Math-1.5B, and Llama3.1-8B, respectively. This robust improvement is attributed to DARO's ability to adaptively modulate the learning signal for each sample based on its difficulty, a capability absent in other methods.

The critical role of dynamic weighting becomes even clearer when contrasted with other techniques. First, incorporating only Dynamic Sampling (DAPO) results in modest and inconsistent gains, highlighting that sampling strategy alone is insufficient. Second, methods employing predefined, static weighting schemes, such as Dr.GRPO and LIPO (proportional to $\sqrt{\mu(1-\mu)}$), fail to generalize and can even be detrimental. For instance, Dr.GRPO lags behind the baseline on all models, and LIPO's performance drops by $-2.1$ points on Llama3.1-8B. This failure underscores the suboptimality of a fixed weighting strategy, which cannot account for model-specific differences or the evolution of model capabilities during training. In conclusion, the results strongly validate that an adaptive, dynamic weighting mechanism is the key factor for achieving superior performance.

\subsection{Training Dynamics}

\begin{figure*}[t]
  \includegraphics[width=\linewidth]{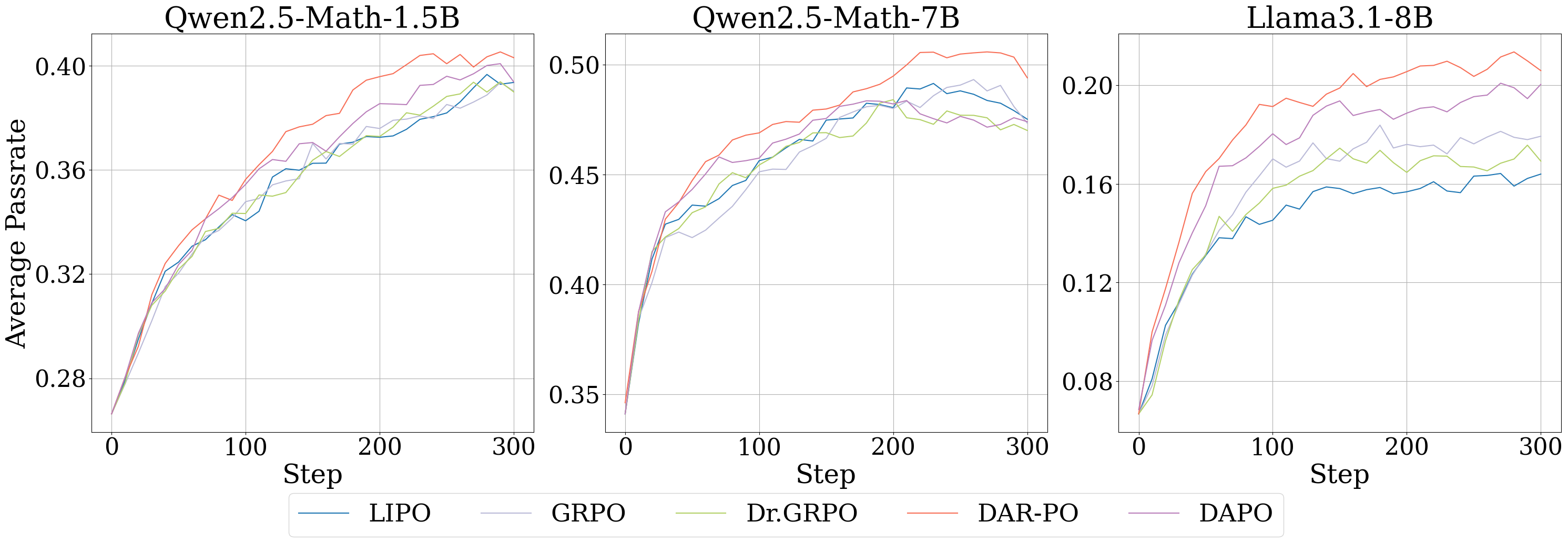}
  \caption{Average passrates of all settings. It is clear that DARO not only converges faster than other methods but also has the highest passrate during the whole training process.}
  \label{fig:passrates}
\end{figure*}

Figure~\ref{fig:train_dynamic} illustrates the dynamics of entropy, response lengths and train rewards throughout the training process. DARO maintains a higher entropy than baselines across the entire training trajectory. This sustained level of policy entropy suggests that our adaptive weighting mechanism successfully mitigates premature convergence, fostering a more robust and continuous exploration of the solution space. Concurrently, the length of the generated responses increases steadily, demonstrating not only a gradually improving reasoning ability but also the model's capacity to develop more complex, multi-step thought processes.

The average accuracy during training is shown in Figure~\ref{fig:passrates}. DARO continuously outperforms other baselines, demonstrating both a faster convergence rate and superior final performance. This robust performance underscores the effectiveness of its dynamic weighting mechanism in alleviating  the loss scaling issue that can hinder methods like GRPO. The efficiency of DARO is particularly evident when training Llama3.1-8B, where it reached a 20\% pass rate using only half the training steps required by DAPO. Furthermore, when training Qwen2.5-Math-7B and Llama3.1-8B, DARO converges at performance levels that are unattainable by the other algorithms, including those employing static weighting strategies like LIPO and Dr.GRPO. This highlights the critical advantage of dynamically adapting the loss weights to the model's evolving state, rather than relying on a fixed, predefined heuristic.

%% file: sections/conclusion.tex
\section{Conclusion}
In this paper, we introduce a unified weighting framework to analyze prominent GRPO-based RLVR algorithms, including GRPO, DAPO, LIPO, and Dr.GRPO. Through this analysis, we identify a fundamental limitation inherent in these methods: the loss scale issue. To address this challenge, we propose Difficulty-Aware Reweighting Policy Optimization (DARO), an algorithm featuring a dynamic weighting mechanism that learns adaptive parameters during training. This enables DARO to accommodate variations in reasoning capabilities across different models and throughout the training stages of a single model. Our extensive experiments demonstrate the effectiveness and robustness of DARO, which consistently outperforms strong baselines by achieving both faster convergence rates and superior final performance.

%% file: sections/limitations.tex
\section{Limitation and Future Work}
DARO is designed to be both straightforward to implement and broadly applicable. To ensure computational efficiency and promote reproducibility, our study utilizes three compact base models on standard benchmarks. The primary limitations of this work are related to its scope. Future research will focus on scaling DARO to larger, state-of-the-art models and extending its application to more general reasoning domains beyond mathematics. Furthermore, we plan to investigate its effectiveness in more complex, agentic settings involving multi-turn interaction and tool use to validate its generality.

%% file: sections/appendix.tex
\newpage
\section*{Appendix}
\appendix
\section{Detailed Mathematical Derivations}

This section provides the detailed steps for the key mathematical derivations presented in the section~\ref{sec:analysis}.

\subsection{Derivation of Advantage Values for Binary Rewards}
\label{appendix:adv}
In our analytical framework, we consider a sub-batch $B_k$ where all samples have an empirical pass rate of $\mu = k/K$. For each sample, this means that out of $K$ responses, $k$ have a reward $r_i=1$ (positive samples) and $K-k$ have a reward $r_i=0$ (negative samples).

\textbf{Mean Reward}: The average reward $\bar{r}$ for the group is:

\begin{equation}
\bar{r} = \frac{\sum_{i=1}^K r_i}{K} = \frac{k \cdot 1 + (K-k) \cdot 0}{K} = \frac{k}{K} = \mu
\end{equation}

\textbf{Standard Deviation}: The variance of the rewards, $\sigma^2$, is:

\begin{equation}
    \begin{split}
\sigma^2 &= \frac{1}{K}\sum_{i=1}^K (r_i - \bar{r})^2 \\
&= \frac{1}{K} \left[ \sum_{i \text{ is pos}} (1-\mu)^2 + \sum_{i \text{ is neg}} (0-\mu)^2 \right] \\
&= \frac{1}{K} \left[ k(1-\mu)^2 + (K-k)(-\mu)^2 \right] \\
&= \frac{k(1-\mu)^2 + (K-k)\mu^2}{K}
    \end{split}
\end{equation}

Substituting $\mu = k/K$:

\begin{equation}
    \begin{split}
\sigma^2 &= \frac{k(1-k/K)^2 + (K-k)(k/K)^2}{K} \\
&= \frac{k\left(\frac{K-k}{K}\right)^2 + (K-k)\frac{k^2}{K^2}}{K} \\
&= \frac{k(K-k)^2 + (K-k)k^2}{K^3} \\
&= \frac{k(K-k)(K-k+k)}{K^3} \\
&= \frac{k(K-k)}{K^2} = \mu(1-\mu)
    \end{split}
\end{equation}

Thus the standard deviation is $\sigma = \sqrt{\mu(1-\mu)}$.

\noindent \textbf{Advantage Values}: The advantage is defined as $A_i = (r_i - \bar{r}) / \sigma$. For a \textbf{positive sample} with a reward of $1$, its advantage $A_+$ is:

\begin{equation}
    \begin{split}
A_+ &= \frac{1-\mu}{\sigma} \\
&= \frac{1-\mu}{\sqrt{\mu(1-\mu)}} = \sqrt{\frac{1-\mu}{\mu}} \\
&= \sqrt{\frac{1-k/K}{k/K}} = \sqrt{\frac{K-k}{k}}
    \end{split}
\end{equation}

For a \textbf{negative sample} with a reward of $0$, its advantage $A_-$ is:

\begin{equation}
    \begin{split}
A_- &= \frac{0-\mu}{\sigma} \\
&= \frac{-\mu}{\sqrt{\mu(1-\mu)}} = -\sqrt{\frac{\mu}{1-\mu}} \\
&= -\sqrt{\frac{k/K}{1-k/K}} = -\sqrt{\frac{k}{K-k}}
    \end{split}
\end{equation}

\subsection{Derivation of The Unified Loss Formulation}
\label{appendix:obj}

We begin with the original form of the loss function of GRPO algorithm:

\begin{equation}
    \begin{split}
    &\mathcal{L}_{GRPO} \\
    =&- \mathbf{E}_{q\sim P(\mathcal{Q}), \{ o_{i} \}^{K}_{i=1}\sim \pi_{\theta_{old}}(\cdot|q)}\bigg[\frac{1}{K}\sum_{i=1}^{K} \frac{1}{\lvert o_{i} \rvert }\sum_{t=1}^{\lvert o_{i} \rvert } \\
    & \min_{}\left( r_{i,t}A_{i,t}, \text{clip}\left( r_{i,t},1-\epsilon,1+\epsilon \right)A_{i,t} \right) \\
    &-\beta \mathbf{D}_{KL}\bigg] \text{where } r_{i,t} = \frac{\pi_{\theta}(o_{i,t}|q,o_{i,<t})}{\pi_{\theta_{old}}(o_{i,t}|q,o_{i,<t})}
    \end{split}
\end{equation}

According to recent works \cite{lipo, yu2025dapo}, the KL loss item $\mathbf{D}_{KL}=\mathbf{D}_{KL}[\pi_{\theta}||\pi_{ref}]$ is useless or even harmful in the RLVR training, and thus we remove it from the original GRPO loss. Meanwhile, the clip operator can be abstracted as a function $f(A,o)$ which takes in one responce and its advantage to yield a value in the range $[0, (1+\epsilon)A]$ when $A>0$ or in the range $[(1-\epsilon)A,0]$. Then $\mathcal{L}_{GRPO}$ can be simpilified to

\begin{equation}
    \begin{split}
\mathcal{L}_{GRPO}=-&\mathbf{E}_{q\sim P(\mathcal{Q}), \{ o_{i} \}^{K}_{i=1}\sim \pi_{\theta_{old}}(\cdot|q)} \\
&\left[\frac{1}{K}\sum_{i=1}^{K} \frac{1}{\lvert o_{i} \rvert } \sum_{t=1}^{\lvert o_{i} \rvert } f(A_{i,t},o_{i,t}) \right]
    \end{split}
\end{equation}

According to the empirical result of \cite{lipo}, the "token-mean" loss aggregation method is better than the "sequence-mean" method used in GRPO in most cases. Considering this fact, we decide to focus on the "token-mean" mode, which changes $\mathcal{L}_{GRPO}$ to the final form:

\begin{equation}
    \begin{split}
        \hat{\mathcal{L}}_{GRPO}= -& \mathbf{E}_{q\sim P(\mathcal{Q}), \{ o_{i} \}^{K}_{i=1}\sim \pi_{\theta_{old}}(\cdot|q)} \\
 & \left[\frac{1}{K} \sum_{i=1}^{K} w_{i} \cdot \frac{\lvert o_i\rvert}{L} f(A_{i},o_{i})\right] \\
 & \text{where } w_{i} = 1
    \end{split}
\end{equation}

For \textbf{DAPO}, the loss function is

\begin{equation}
    \begin{split}
&\mathcal{L}_{DAPO} \\
=-&\mathbf{E}_{q\sim P(\mathcal{Q}), \{ o_{i} \}^{K}_{i=1}\sim \pi_{\theta_{old}}(\cdot|q),\mu\ne0,1} \\
&\left[\frac{1}{KL}\sum_{i=1}^{K} \sum_{t=1}^{\lvert o_{i} \rvert } f(A_{i,t},o_{i,t})\right] \\
=-& \mathbf{E}_{q\sim P(\mathcal{Q}), \{ o_{i} \}^{K}_{i=1}\sim \pi_{\theta_{old}}(\cdot|q)} \\
&\left[\frac{1}{K}\sum_{i=1}^{K} w_{i}\cdot \frac{\lvert o_{i} \rvert}{L} f(A_{i}, o_{i})\right] \\
&\text{ where } w_{i}=\mathbb{I}(0<\mu<1)
    \end{split}
    \label{eq:dapo_loss}
\end{equation}

For \textbf{LIPO}, by utilizing the linearity of clipping function $f$ with respect to the variable $A$, we can obtain

\begin{equation}
    \begin{split}
        &f(\frac{r_i-\mu}{\hat{\sigma}}, \cdot)\\
        =&f(\frac{\sigma}{\hat{\sigma}}\frac{r_i-\mu}{\sigma}, \cdot) =\frac{\sigma}{\hat{\sigma}} f(\frac{r_i-\mu}{\sigma}, \cdot) 
    \end{split}
    \label{eq:lipo_linear}
\end{equation}

Substitute the Equation~\ref{eq:lipo_linear} to Equation~\ref{eq:dapo_loss}, then we can get

\begin{equation}
    \begin{split}
&\mathcal{L}_{LIPO} \\
=-&\mathbf{E}_{q\sim P(\mathcal{Q}), \{ o_{i} \}^{K}_{i=1}\sim \pi_{\theta_{old}}(\cdot|q)} \\
& \left[\frac{1}{KL}\sum_{i=1}^{K} \lvert o_{i} \rvert f(\frac{r_{i}-\mu}{\hat{\sigma}},o_{i})\right] \\
=-&\mathbf{E}_{q\sim P(\mathcal{Q}), \{ o_{i} \}^{K}_{i=1}\sim \pi_{\theta_{old}}(\cdot|q)} \\
& \left[\frac{1}{K}\sum_{i=1}^{K} \frac{\sigma}{\hat{\sigma}} \cdot\frac{\lvert o_{i} \rvert}{L} f(\frac{r_{i}-\mu}{\sigma}, o_{i})\right] \\
=-&\mathbf{E}_{q\sim P(\mathcal{Q}), \{ o_{i} \}^{K}_{i=1}\sim \pi_{\theta_{old}}(\cdot|q)} \\
& \left[\frac{1}{K}\sum_{i=1}^{K} w_{i}\cdot \frac{\lvert o_{i} \rvert}{L} f(A_{i}, o_{i}) \right] \text{ where } w_{i} = \frac{\sigma}{\hat{\sigma}}     
    \end{split}
\end{equation}

For \textbf{Dr.GRPO}, using the linearity of $f$ again, the loss function is

\begin{equation}
    \begin{split}
&\mathcal{L}_{Dr.GRPO} \\
=-&\mathbf{E}_{q\sim P(\mathcal{Q}), \{ o_{i} \}^{K}_{i=1}\sim \pi_{\theta_{old}}(\cdot|q)} \\
&\left[\frac{1}{K}\sum_{i=1}^{K} \lvert o_{i} \rvert f(r_{i}-\mu,o_{i})\right] \\
=-&\mathbf{E}_{q\sim P(\mathcal{Q}), \{ o_{i} \}^{K}_{i=1}\sim \pi_{\theta_{old}}(\cdot|q)} \\
&\left[\frac{1}{K}\sum_{i=1}^{K} L\sigma \cdot \frac{\lvert o_{i} \rvert}{L} f(\frac{r_{i}-\mu}{\sigma}, o_{i})\right] \\
=-&\mathbf{E}_{q\sim P(\mathcal{Q}), \{ o_{i} \}^{K}_{i=1}\sim \pi_{\theta_{old}}(\cdot|q)} \\
&\left[\frac{1}{K}\sum_{i=1}^{K} w_{i}\cdot \frac{\lvert o_{i} \rvert}{L} f(A_{i}, o_{i}) \right] \text{ where } w_{i} = L\sigma
    \end{split}
\end{equation}

To conclude, the loss of the three algorithms above can be unified to be the same form:

\begin{equation}
    \begin{split}
\mathcal{L}_{BASE}= -& \mathbf{E}_{q\sim P(\mathcal{Q}), \{ o_{i} \}^{K}_{i=1}\sim \pi_{\theta_{old}}(\cdot|q)} \\
&\left[\frac{1}{K} \sum_{i=1}^{K} w_{i} \cdot \frac{\lvert o_{i} \rvert }{L} A_{i}f(o_{i})\right]
    \end{split}
\end{equation}

\subsection{Proof of The Approximation}
\label{appendix:lemma1}

Let the full batch be $B=\bigcup_{\mu} G_{\mu}$, where group $G_\mu$ with $\mu=k/K$ is composed of samples with $k$ positive responses in $B$. We first approximate the sub-loss $\mathcal{L}_{BASE}(G_\mu)$ for a group $G_\mu$. The original form is:

\begin{equation}
    \begin{split}
    &\mathcal{L}_{BASE}(G_\mu)=-\frac{1}{nK} [\mathcal{L}_{+}(G_\mu) + \mathcal{L}_{-}(G_\mu)] \\
    &\text{where }\mathcal{L}_{\pm}(G_\mu)=\sum_{o_i^{(j)} \text{ is pos/neg}} \frac{\lvert o_i^{(j)}\rvert}{L} f(A_{\pm}, o_{i}^{(j)}). \\
    \end{split}
\end{equation}

For $\mathcal{L}_{+}(G_\mu)$: This term is a weighted sum of $N_+ = kn$ random variables $f(A_{+}, o_{i}^{(j)})$. Since $f(A_+,\cdot) \in [0, (1+\epsilon)A_+]$, which has a width of $(1+\epsilon)A_+$. It can be assumed that $\mathbf{E}[f(A_+, \cdot)]=A_+$, since in the practical training process, the model is updated to a very small extent at each step, which leads to $r(\theta)=\frac{\pi_{\theta}(o_{i,t})}{\pi_{\theta_{old}}(o_{i,t})}\approx 1$. The expectation of $\mathcal{L}_{+}(G_\mu)$ is:

\begin{equation}
    \begin{split}
\mathbf{E}[\mathcal{L}_{+}(G_\mu)]         &=\sum_{o_i^{(j)} \text{ is pos}} \frac{\lvert o_i^{(j)}\rvert}{L} \mathbf{E}\left[f(A_{+}, o_{i}^{(j)})\right]  \\
&= \sum_{o_i^{(j)} \text{ is pos}}\frac{\lvert o_i^{(j)}\rvert}{L}\cdot\sum_{o_i^{(j)} \text{ is pos}} A_+ \\
& = \frac{L_+}{L}knA_+ = \frac{L_+}{L}n\sqrt{k(K-k)}\\
\text{where } & L_+ = \sum_{o_i^{(j)} \text{ is pos}}\lvert o_i^{(j)}\rvert
    \end{split}
\end{equation}

Since $\mathcal{L}_+(G_\mu)$ is composed of many stochastic variables, it is natural to use Hoeffding's Inequality \cite{hoeffdings_inequality} to analyze its distribution.

\textbf{Hoeffding's Inequality}: Let $X_1, \dots, X_N$ be independent random variables such that $X_i \in [a_i, b_i]$. Their sum $S_N = \sum_{i=1}^N X_i$ satisfies the following inequality for any $\delta > 0$:
\begin{equation}
    \begin{split}
        &\mathbf{P}(\lvert S_N - \mathbf{E}[S_N] \rvert \geq \delta) \\
        &\quad\leq 2\exp\left(-\frac{2\delta^2}{\sum_{i=1}^N (b_i - a_i)^2} \right)
    \end{split}
\end{equation}

Applying Hoeffding's inequality to $\mathcal{L}_+(G_\mu)$, we get:

\begin{equation}
    \begin{split}
&\mathbf{P}(\lvert \mathcal{L}_{+}(G_\mu)-\mathbf{E}[\mathcal{L}_{+}(G_\mu)]  \rvert\geq \delta) \\
&\leq 2\exp\left( -\frac{2\delta^2}{\sum_{i=1}^{kn} ((1+\epsilon) A_+)^2} \right) \\
&=  2\exp\left( -\frac{2\delta^2}{n(K-k)(1+\epsilon)^2} \right)
    \end{split}
    \label{eq:LGk+}
\end{equation}

Similarly, for $\mathcal{L}_{-}(G_\mu)$, apply the Hoeffding's inequality and we can get

\begin{equation}
    \begin{split}
&\mathbf{P}(\lvert \mathcal{L}_{-}(G_\mu)-\mathbf{E}[\mathcal{L}_{-}(G_\mu)] \rvert \geq \delta) \\
&\leq 2\exp\left( -\frac{2\delta^2}{\sum_{i=1}^{(K-k)n} (\lvert(1-\epsilon) A_-\rvert)^2} \right)\\
&= 2\exp\left( -\frac{2\delta^2}{kn(1-\epsilon)^2} \right)
    \end{split}
    \label{eq:LGK-}
\end{equation}

Additionally, the above results still hold in the case that \textit{clip-higher} trick is used. In such case, the only difference is that the $\epsilon$ in Equation~\ref{eq:LGk+} is replaced with a larger value $\epsilon_{\text{high}}$, which leads to a negligible change.

Now return to $\mathcal{L}_{BASE}(G_\mu)$. According to Equation~\ref{eq:LGk+} and ~\ref{eq:LGK-}, we know that the probability of $\mathcal{L}_{\pm}(G_\mu)$ deviating from their means $\mathbf{E}[\mathcal{L}_{\pm}]$ decreases square-exponentially concentrated around the mean value of them, thus an approximation can be applied to the original form of $\mathcal{L}_{BASE}(G_\mu)$:

\begin{equation}
    \begin{split}
&\mathcal{L}_{BASE}(G_{\mu}) \\
= & -\frac{1}{nK} [\mathcal{L}_{+}(G_\mu) + \mathcal{L}_{-}(G_\mu)] \\
\approx & -\frac{1}{nK}\left(\frac{L_+}{L}n\sqrt{k(K-k)}-\frac{L_-}{L}n\sqrt{k(K-k)}\right) \\
= & \frac{\sqrt{k(K-k)}}{KL}(L_+-L_-)\\
= & \frac{L_+ - L_-}{L} \sqrt{\mu(1-\mu)} \\
= & \frac{\sum_{o\text{ is pos}} \lvert o\rvert - \sum_{o\text{ is neg}} \lvert o\rvert}{L} \sqrt{\mu(1-\mu)}
    \end{split}
\end{equation} $\square$

\section{Detailed Parameters}
\label{appendix:detailed_params}

Table~\ref{tab:parameters} details the training parameters of DARO and the baselines. In addition, the loss aggregation mode is \textit{token-mean} except Dr.GRPO, whose loss aggregation mode is \textit{seq-mean-token-sum-norm}.

\begin{table*}[h!]
    \centering
    \caption{Detailed Parameters for DARO and four baselines.}
    \begin{tabular}{l l c c c c c c}
         \toprule
         \textbf{Category} & \textbf{Parameter} & \textbf{GRPO} & \textbf{LIPO} & \textbf{Dr.GRPO} & \textbf{DAPO} & \textbf{DARO} \\
         \midrule
         \multirow{5}{*}{\textbf{General}}
         & use kl in reward & False & False & False & False & False \\
         & use kl loss & False & False & False & False & False \\
         & tensor parallel size & 1 & 1 & 1 & 1 & 1 \\
         & total training steps & 300 & 300 & 300 & 300 & 300 \\
         & test frequency & 10 & 10 & 10 & 10 & 10 \\
         \midrule
         \multirow{2}{*}{\textbf{Clip Higher}}
         & clip-high & 0.28 & 0.28 & 0.28 & 0.28 & 0.28 \\
         & clip-low & 0.2 & 0.2 & 0.2 & 0.2 & 0.2 \\
         \midrule
         \multirow{4}{*}{\textbf{Training}}
         & train batch size & 128 & 128 & 128 & 128 & 128 \\
         & mini batch size & 64 & 64 & 64 & 64 & 64 \\
         & gen batch size & - & - & - & 384 & 384 \\
         & gradient clip & 0.5 & 0.5 & 0.5 & 0.5 & 0.5 \\
         & warmup steps & 0 & 0 & 0 & 0 & 0 \\
         \midrule
         \multirow{5}{*}{\textbf{Rollout}}
         & max prompt length & 1024 & 1024 & 1024 & 1024 & 1024 \\
         & max responce length & 8192 & 8192 & 8192 & 8192 & 8192 \\
         & n & 8 & 8 & 8 & 8 & 8 \\
         & do sample & False & False & False & False & False \\
         & temperature & 1.0 & 1.0 & 1.0 & 1.0 & 1.0 \\
         & top_p & 1.0 & 1.0 & 1.0 & 1.0 & 1.0 \\
         \midrule
         \multirow{1}{*}{\textbf{Filter Group}}
         & enable & False & False & False & True & True \\
         \midrule
         \multirow{1}{*}{\textbf{Optimizer}}
         & lr & $1\mathrm{e}{-6}$ & $1\mathrm{e}{-6}$ & $1\mathrm{e}{-6}$ & $1\mathrm{e}{-6}$ & $1\mathrm{e}{-6}$ \\
         \midrule
         \multirow{3}{*}{\textbf{Validation}}
         & do sample & True & True & True & True & True \\
         & temperature & 0.6 & 0.6 & 0.6 & 0.6 & 0.6 \\
         & top_p & 1.0 & 1.0 & 1.0 & 1.0 & 1.0 \\
         \midrule
         \multicolumn{7}{c}{\textit{Llama3.1-8B Specific}} \\
         \midrule
         \multirow{1}{*}{\textbf{Training}}
         & max responce length & 2048 & 2048 & 2048 & 2048 & 2048 \\
         \bottomrule
    \end{tabular}
    \label{tab:parameters}
\end{table*}

\section{Detailed Training Results}
\label{appendix:detailed_results}

\begin{figure*}[ht]
  \includegraphics[width=\linewidth]{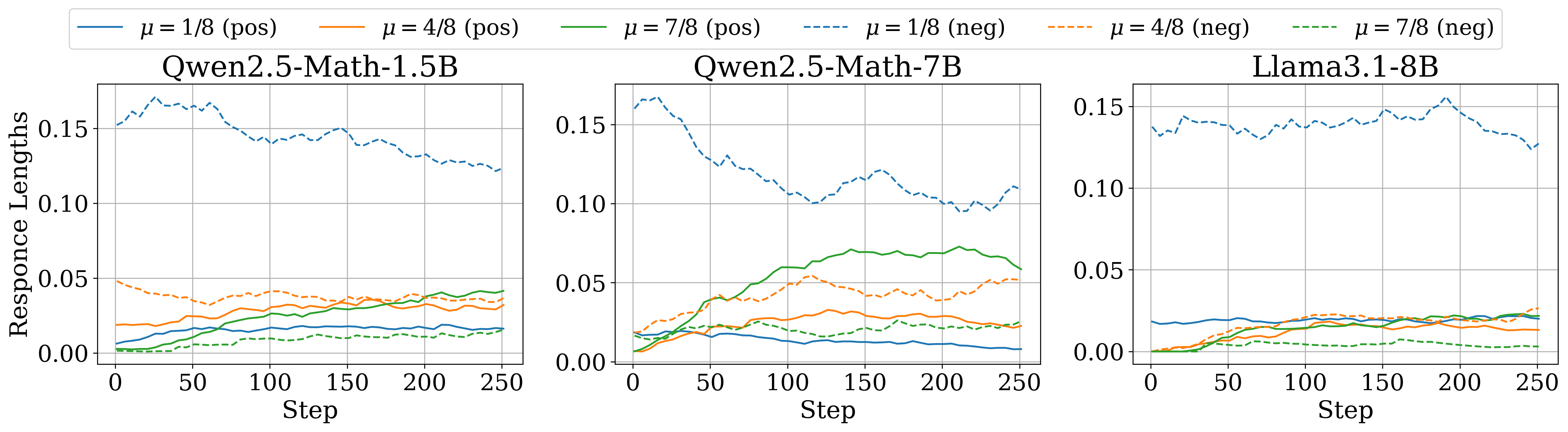}
  \caption{\textbf{Normalized Response Lengths} of three considered models in the GRPO training process. The responce lengths are normalized by dividing the factor of total responce length across the batch $B$ at each step. The solid and dashed lines represents the responce lengths of samples with reward $1$ (positive) and $0$ (negative), respectively.}
  \label{fig:responce_lengths_combined}
\end{figure*}

In this section, we show detailed training results throughout the entire training process. Figure~\ref{fig:responce_lengths_combined} illustrates the correlation between responce lengths and difficulty levels of samples, which is asserted in Section~\ref{sec:loss_scale_issue}. Table~\ref{tab:step_results} indicates that under the same number of steps for all three base model, \textbf{DARO always reaches an optimal average accuracy with no exception}. At the same time, DARO maintains the optimal or sub-optimal on each benchmark.

\begin{table*}[h!]
    \centering
    \caption{\textbf{Validation accuracy (\%) at step 100, 200 and 300}. The best result in each column for a given model is highlighted in \textbf{\color{RoyalBlue}blue} except the "Average" column with \textbf{\color{WildStrawberry}red}. When the value in \textbf{DARO} row is not the best, it is underlined if sub-optimal. The proposed DARO method is marked with a gray background.}
    \sisetup{table-format=2.1, detect-weight} 
    \resizebox{0.8\linewidth}{!}{%
    \begin{tabular}{p{40pt} S[table-format=2.1] S[table-format=2.1] S[table-format=2.1] S[table-format=1.1] S[table-format=2.1] S[table-format=2.1] S[table-format=2.1] S[table-format=2.1, detect-weight, table-text-alignment=center, table-number-alignment=center]}
        \toprule
         & {\bf \makecell{MATH500\\mean@1}} & {\bf \makecell{AMC23\\mean@32}} & {\bf \makecell{AIME24\\mean@32}} & {\bf \makecell{AIME25\\mean@32}} & {\bf \makecell{Olympiad\\mean@1}} & {\bf \makecell{Minerva\\mean@1}} & {\bf \makecell{gsm8k\\mean@1}} & {\textbf{Average}} \\
        \midrule
        \multicolumn{9}{c}{\textbf{Validation Accuracy (\%) at step 100}} \\ 
        \midrule
        \multicolumn{9}{c}{\textit{Llama-3.1-8B}} \\
        \midrule
        GRPO    & 24.4 &  8.2 & \highlightbest{0.6}{0.6} & 0.0 & 7.3  & \highlightbest{19.1}{19.1} & 61.6 & 17.3 \\
        LIPO    & 20.6 &  8.4 & 0.3  & 0.2 & 6.1  & 11.0 & 55.5 & 14.6 \\
        Dr.GRPO & 21.6 &  9.5 & 0.1  & 0.0 & 6.4  & 16.9 & 58.1 & 16.1 \\
        DAPO    & 26.2 & \highlightbest{11.6}{11.6} & 0.4  & \highlightbest{0.5}{0.5} & 7.1  & 15.8 & 66.1 & 18.3 \\
        \rowcolor{Gray}
        DARO  & \highlightbest{30.0}{30.0} & 9.8 & \highlightbest{0.6}{0.6}  & \underline{0.3} & \highlightbest{7.7}{7.7} & 14.7 & \highlightbest{70.6}{70.6} & \bfseries {\color{WildStrawberry}{19.1}} \\
        \midrule
        \multicolumn{9}{c}{\textit{Qwen2.5-Math-1.5B}} \\
        \midrule
        GRPO    & \highlightbest{65.6}{65.6} & 41.6 & 9.4  & 4.4 & 29.4 & 23.9 & 71.1 & 35.0 \\
        LIPO    & 63.6 & 40.8 & 8.9 & 4.1 & 29.4 & 20.2 & 70.7 & 33.9 \\
        Dr.GRPO & 62.6 & 40.8 & \highlightbest{10.0}{10.0} & 4.1 & 29.1 & 25.7 & 67.9 & 34.3 \\
        DAPO    & 64.2 & \highlightbest{43.0}{43.0} & 9.7  & \highlightbest{4.5}{4.5} & 28.6 & 27.2 & 72.5 & 35.7 \\
        \rowcolor{Gray}
        DARO  & 62.6 & \underline{42.2} & 9.6 & \underline{4.4} & \highlightbest{33.1}{33.1} & \highlightbest{27.2}{27.2} & \highlightbest{72.8}{72.8} & \bfseries {\color{WildStrawberry}{36.0}} \\
        \midrule
        \multicolumn{9}{c}{\textit{Qwen2.5-Math-7B}} \\
        \midrule
        GRPO    & 77.4 & 56.3 & 17.6 & 8.9 & 39.1 & 34.9 & 84.2 & 45.5 \\
        LIPO    & 76.0 & 57.6 & 18.0 & \highlightbest{10.0}{10.0} & 41.0 & 33.1 & \highlightbest{86.4}{86.4} & 46.0 \\
        Dr.GRPO & 77.4 & \highlightbest{57.9}{57.9} & 16.8 & 9.0 & \highlightbest{41.2}{41.2} & 32.7 & 84.9 & 45.7 \\
        DAPO    & 77.2 & 56.6 & 18.8 & 8.6 & 40.6 & 32.7 & 86.1 & 45.8 \\
        \rowcolor{Gray}
        DARO  & \highlightbest{79.6}{79.6} & \underline{57.6} & \highlightbest{19.7}{19.7} & \underline{9.1} & 39.8 & \highlightbest{37.9}{37.9} & 85.0 & \bfseries {\color{WildStrawberry}{46.9}} \\
        \midrule
        \multicolumn{9}{c}{\textbf{Validation Accuracy (\%) at step 200}} \\
        \midrule
        \multicolumn{9}{c}{\textit{Llama-3.1-8B}} \\
        \midrule
        GRPO    & 24.0 &  11.9 & 0.5 & \highlightbest{0.3}{0.3} & 6.5 & 15.1 & 65.4 & 17.7 \\
        LIPO    & 21.2 &  9.7 & 0.0  & 0.0 & 6.7  & 16.5 & 55.9 & 15.7 \\
        Dr.GRPO & 21.4 &  9.1 & 0.3  & 0.1 & 5.5  & 17.3 & 60.3 & 16.7 \\
        DAPO    & 27.8 & 10.9 & \highlightbest{0.9}{0.9} & \highlightbest{0.3}{0.3} & 7.1 & 18.8 & 67.1 & 19.0 \\
        \rowcolor{Gray}
        DARO  & \highlightbest{31.0}{31.0} & \highlightbest{14.1}{14.1} & 0.2  & \underline{0.2} & \highlightbest{8.9}{8.9} & \highlightbest{19.1}{19.1} & \highlightbest{71.0}{71.0} & \bfseries {\color{WildStrawberry}{20.7}} \\
        \midrule
        \multicolumn{9}{c}{\textit{Qwen2.5-Math-1.5B}} \\
        \midrule
        GRPO    & 66.6 & 44.4 & 10.6 & 5.8 & 31.1 & 27.6 & 76.8 & 37.6 \\
        LIPO    & 66.6 & 45.0 & 9.4 & 6.0 & 30.5 & 26.8 & 76.3 & 37.2 \\
        Dr.GRPO & 67.8 & 42.8 & 10.1 & \highlightbest{6.7}{6.7} & 31.6 & 26.5 & 75.4 & 37.3 \\ 
        DAPO    & 69.2 & 44.9 & 11.4 & 4.7 & 33.8 & \highlightbest{28.3}{28.3} & 78.5 & 38.7 \\
        \rowcolor{Gray}
        DARO  & \highlightbest{71.2}{71.2} & \highlightbest{46.1}{46.1} & \highlightbest{11.7}{11.7} & \underline{6.3} & \highlightbest{34.3}{34.3} & 26.5 & \highlightbest{81.4}{81.4} & \bfseries {\color{WildStrawberry}{39.6}} \\
        \midrule
        \multicolumn{9}{c}{\textit{Qwen2.5-Math-7B}} \\
        \midrule
        GRPO    & 79.6 & 59.8 & 17.0 & 12.5 & 42.5 & 36.0 & 88.2 & 47.9 \\
        LIPO    & 80.0 & 58.9 & 17.9 & 10.4 & 43.1 & 37.5 & 88.1 & 48.0 \\
        Dr.GRPO & 79.8 & 60.3 & \highlightbest{19.4}{19.4} & \highlightbest{13.8}{13.8} & 43.1 & 35.3 & 87.7 & 48.5 \\
        DAPO    & 81.6 & 58.2 & 15.7 & 12.5 & 44.0 & 36.4 & 88.7 & 48.2 \\
        \rowcolor{Gray}
        DARO  & \highlightbest{82.0}{82.0} & \highlightbest{61.9}{61.9} & 17.8 & \underline{12.5} & \highlightbest{44.9}{44.9} & \highlightbest{38.2}{38.2} & \highlightbest{90.2}{90.2} & \bfseries {\color{WildStrawberry}{49.6}} \\
        \midrule
        \multicolumn{9}{c}{\textbf{Validation Accuracy (\%) at step 300}} \\
        \midrule
        \multicolumn{9}{c}{\textit{Llama-3.1-8B}} \\
        \midrule
        GRPO    & 24.4 & \highlightbest{11.4}{11.4} & 0.3 & \highlightbest{0.3}{0.3} & 7.3 & 15.1 & 67.2 & 18.0 \\
        LIPO    & 18.2 & 8.1 & 0.2 & 0.0 & 5.8 & 19.9 & 63.2 & 16.5 \\
        Dr.GRPO & 21.4 & 7.9 & 0.0 & \highlightbest{0.3}{0.3} & 6.1 & 19.5 & 61.3 & 16.6 \\
        DAPO    & 32.0 & 10.7 & 0.9 & 0.0 & \highlightbest{9.2}{9.2} & \highlightbest{20.2}{20.2} & 69.0 & 20.3 \\
        \rowcolor{Gray}
        DARO  & \highlightbest{32.4}{32.4} & 10.3 & \highlightbest{1.8}{1.8} & \underline{0.2} & \underline{8.3} & 17.3 & \highlightbest{72.7}{72.7} & \bfseries {\color{WildStrawberry}{20.4}} \\
        \midrule
        \multicolumn{9}{c}{\textit{Qwen2.5-Math-1.5B}} \\
        \midrule
        GRPO    & 71.6 & 45.3 & 8.4 & 6.9 & 33.3 & 27.6 & 79.2 & 38.9 \\
        LIPO    & 71.4 & \highlightbest{47.7}{47.7} & 10.1 & 5.3 & 33.6 & 27.9 & 79.6 & 39.4 \\
        Dr.GRPO & 67.8 & 45.1 & \highlightbest{11.0}{11.0} & 6.7 & 34.0 & 26.5 & 80.8 & 38.8 \\
        DAPO    & 70.6 & 43.8 & 9.4 & 5.7 & 35.3 & 26.8 & \highlightbest{82.0}{82.0} & 39.1 \\
        \rowcolor{Gray}
        DARO  & \highlightbest{71.8}{71.8} & \underline{47.0} & \underline{10.6} & \highlightbest{7.4}{7.4} & \highlightbest{36.2}{36.2} & \highlightbest{29.3}{29.3} & 79.3 & \bfseries {\color{WildStrawberry}{40.2}} \\
        \midrule
        \multicolumn{9}{c}{\textit{Qwen2.5-Math-7B}} \\
        \midrule
        GRPO    & 79.2 & 58.6 & 14.3 & \highlightbest{13.4}{13.4} & \highlightbest{44.4}{44.4} & 33.8 & 85.4 & 47.0 \\
        LIPO    & 81.4 & 56.6 & 13.1 & 11.7 & 42.8 & \highlightbest{36.8}{36.8} & 89.1 & 47.4 \\
        Dr.GRPO & 80.2 & 56.3 & 11.9 & 11.3 & 42.8 & 36.0 & 89.8 & 46.9 \\
        DAPO    & 78.8 & 58.5 & 14.5 & 11.7 & 41.6 & 36.4 & 89.8 & 47.3 \\
        \rowcolor{Gray}
        DARO  & \highlightbest{81.8}{81.8} & \highlightbest{61.3}{61.3} & \highlightbest{17.3}{17.3} & \underline{12.0} & \underline{42.8} & 36.0 & \highlightbest{91.8}{91.8} & \bfseries {\color{WildStrawberry}{49.0}} \\
        \bottomrule
    \end{tabular}%
    \label{tab:step_results}
    }
\end{table*}